\documentclass[letterpaper]{article} 
\usepackage{aaai23}  
\usepackage{times}  
\usepackage{helvet}  
\usepackage{courier}  
\usepackage[hyphens]{url}  
\usepackage{graphicx} 
\usepackage{natbib}  
\usepackage{caption} 
\usepackage{algorithm}
\usepackage{algorithmic}
\usepackage{xcolor}  
\usepackage{amsmath}
\usepackage{cleveref}
\usepackage{graphicx}
\usepackage{booktabs} 
\usepackage{multirow}
\usepackage{makecell}
\usepackage{bm}
\usepackage{amsfonts} 
\usepackage{pifont}
\newcommand{\cmark}{\ding{51}}%
\newcommand{\xmark}{\ding{55}}%
\def\x{$\times$}
\newcommand{\ie}{\textit{i}.\textit{e}.}
\newcommand{\eg}{\textit{e}.\textit{g}.}
\definecolor{demphcolor}{RGB}{144,144,144}

\usepackage{colortbl}

\definecolor{baselinecolor}{gray}{.9}
\newcommand{\baseline}[1]{\cellcolor{baselinecolor}{#1}}
\newlength\savewidth
\newcommand{\vs}{\textit{v}.\textit{s}.}
\definecolor{mygray}{gray}{0.95}
\usepackage{listings}
\DeclareFixedFont{\ttb}{T1}{txtt}{bx}{n}{6} 
\DeclareFixedFont{\ttm}{T1}{txtt}{m}{n}{6}  

\definecolor{deepblue}{rgb}{0,0,0.5}
\definecolor{deepred}{rgb}{0.6,0,0}
\definecolor{deepgreen}{rgb}{0,0.5,0}

\definecolor{codegreen}{rgb}{0,0.6,0}
\definecolor{codegray}{rgb}{0.5,0.5,0.5}
\definecolor{codepurple}{rgb}{0.58,0,0.82}
\definecolor{backcolour}{rgb}{0.95,0.95,0.92}

\definecolor{codeblue}{rgb}{0.25,0.5,0.5}
\usepackage{newfloat}
\usepackage{listings}
\DeclareCaptionStyle{ruled}{labelfont=normalfont,labelsep=colon,strut=off} 
\floatstyle{ruled}
\newfloat{listing}{tb}{lst}{}
\floatname{listing}{Listing}
\title{Revisiting Classifier: Transferring Vision-Language Models for Video Recognition}
\author{%
Wenhao Wu\textsuperscript{\rm 1},
Zhun Sun\textsuperscript{\rm2},
Wanli Ouyang\textsuperscript{\rm3}\thanks{Corresponding author.}
}
\affiliations{
    \textsuperscript{\rm 1}The University of Sydney, NSW, Australia\\
    \textsuperscript{\rm 2}Baidu Inc., Beijing, China\\
    \textsuperscript{\rm 3}Shanghai Artificial Intelligence Laboratory, Shanghai, China \\
    whwu.ucas@gmail.com, sunzhun@baidu.com, wanli.ouyang@sydney.edu.au

}

\usepackage{bibentry}

\begin{document}

\maketitle
\begin{abstract}

Transferring knowledge from task-agnostic pre-trained deep models for downstream tasks is an important topic in computer vision research.
Along with the growth of computational capacity, we now have open-source vision-language pre-trained models in large scales of the model architecture and amount of data.
In this study, we focus on transferring knowledge for video classification tasks.
Conventional methods randomly initialize the linear classifier head for vision classification, but they leave the usage of the text encoder for downstream visual recognition tasks undiscovered. 
In this paper, we revise the role of the linear classifier and replace the classifier with different knowledge from the pre-trained model. We utilize the well-pre-trained language model to generate a good semantic target for efficient transferring learning.
The empirical study shows that our method improves both the performance and the training speed of video classification, with a negligible change in the model.
Our simple yet effective tuning paradigm achieves state-of-the-art performance and efficient training on various video recognition scenarios, \ie, zero-shot, few-shot, and general recognition.
In particular, our paradigm achieves the state-of-the-art accuracy of 87.8\% on Kinetics-400, and also surpasses previous methods by 20$\sim$50\% absolute top-1 accuracy under zero-shot, few-shot settings on five video datasets.
Code and models are available at \url{https://github.com/whwu95/Text4Vis}.
\end{abstract}

\section{Introduction}
\label{sec:intro}
Pre-training a task-agnostic model using large-scale general datasets and then transferring its learning feature representations to downstream tasks is a paradigm in many computer vision applications. 
While in the last decade, the convolutional-based models that are optimized on the ImageNet~\cite{deng2009imagenet} dataset with a supervised style dominated this field. Owing to the dramatically increasing computational capacity, now we can train models that have several magnitude more model parameters and FLOPs on various image and even video datasets in either supervised~\cite{JFT300M} or self-supervised~\cite{he2020momentum,huang2021ascnet,fang2022mamico} style.
Recently, contrastive-based vision-language pre-training~\cite{CLIP} manifest their superior capabilities in improving downstream tasks performance such as classification~\cite{CLIP}, captioning~\cite{mokady2021clipcap}, image generation~\cite{DALLE}, to name a few. These models are powerful for two reasons: i) the employed large-scale weakly-related datasets provide rich semantics and diverse representations of concepts; ii) the representation vectors of images and texts are roughly aligned in the semantic embedding space.
However, the most common approach to using these models is fine-tuning the visual encoder on specific tasks. Although the rich semantics and diverse representations of concepts benefit the downstream tasks, the usage of the textual encoder is still left overlooked.

\begin{figure}
\begin{center}
\includegraphics[width=0.48\textwidth]{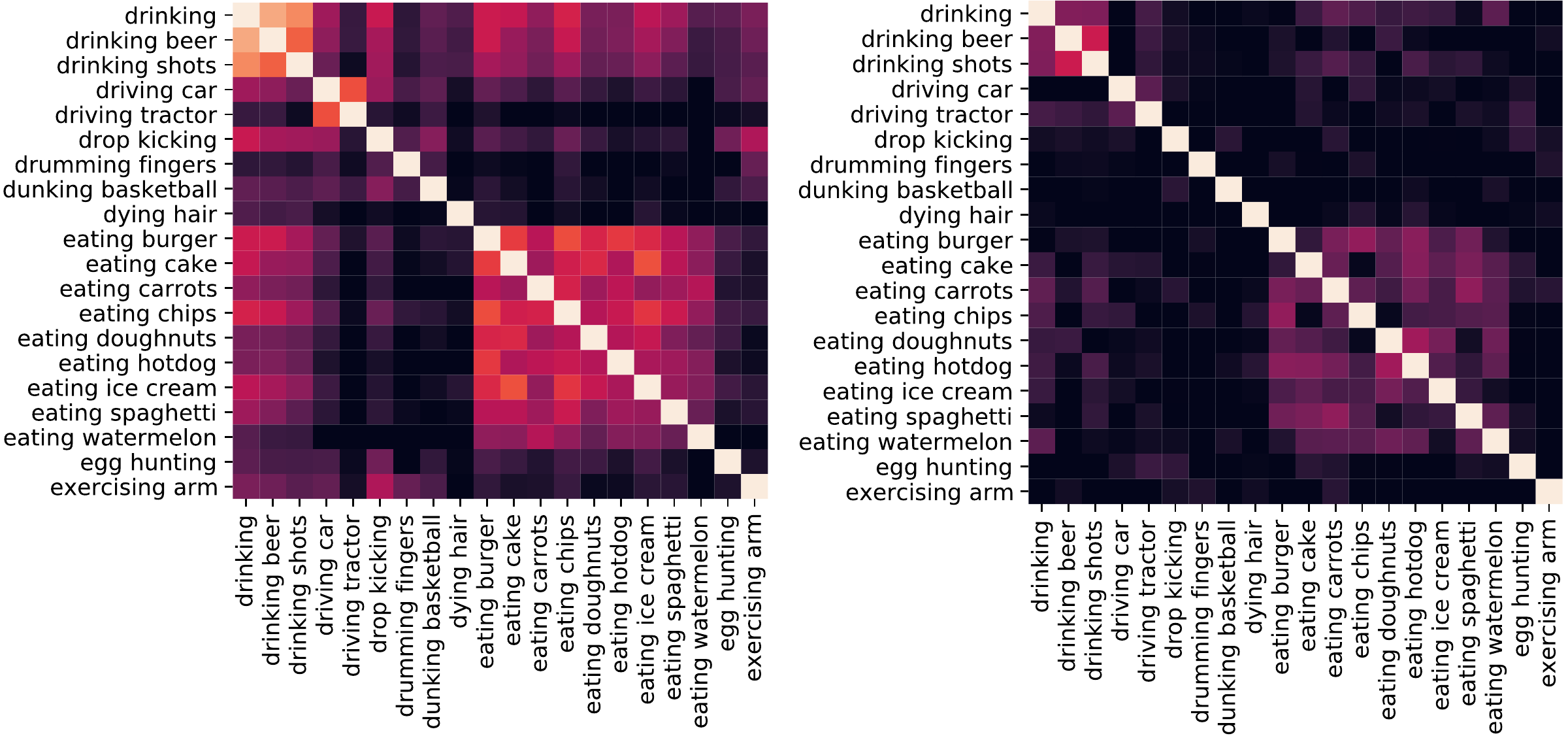}
\end{center}
\caption{Inter-class correlation maps of ``embeddings of class labels'' for 20 categories on Kinetics-400.
\textbf{Left:} The extracted textual vectors of class labels, \textbf{Right:} The ``embeddings'' from learned classifier. The color thresholds are adjusted for a better view. Please zoom in for the best view.}
\label{fig:intro_heatmap}
\end{figure}

In this study, we aim to improve the transferability of such vision-language pre-training models for downstream classification tasks, with the help of their textual encoders. Our motivation comes from the semantic similarity among the ground-truth labels. To demonstrate this, we employ the Kinetics video recognition dataset~\cite{kay2017kinetics} for the analysis. We extract the embedded textual vectors of class labels using the textual encoder of CLIP. We then calculate the correlation between the embedded textual vectors. 
The plot is shown on the left of \Cref{fig:intro_heatmap}. Not surprisingly, the extracted textual vectors of class labels exhibit certain inter-class correlations since part of them include the same verbs in their labels, \eg, \textit{playing} \textless\textit{something}\textgreater. Meanwhile, the labels with different verbs show a negligible inter-class correlation, \eg, \textit{drinking} and \textit{driving}.

\begin{figure*}[t]
\begin{center}
\includegraphics[width=0.98\linewidth]{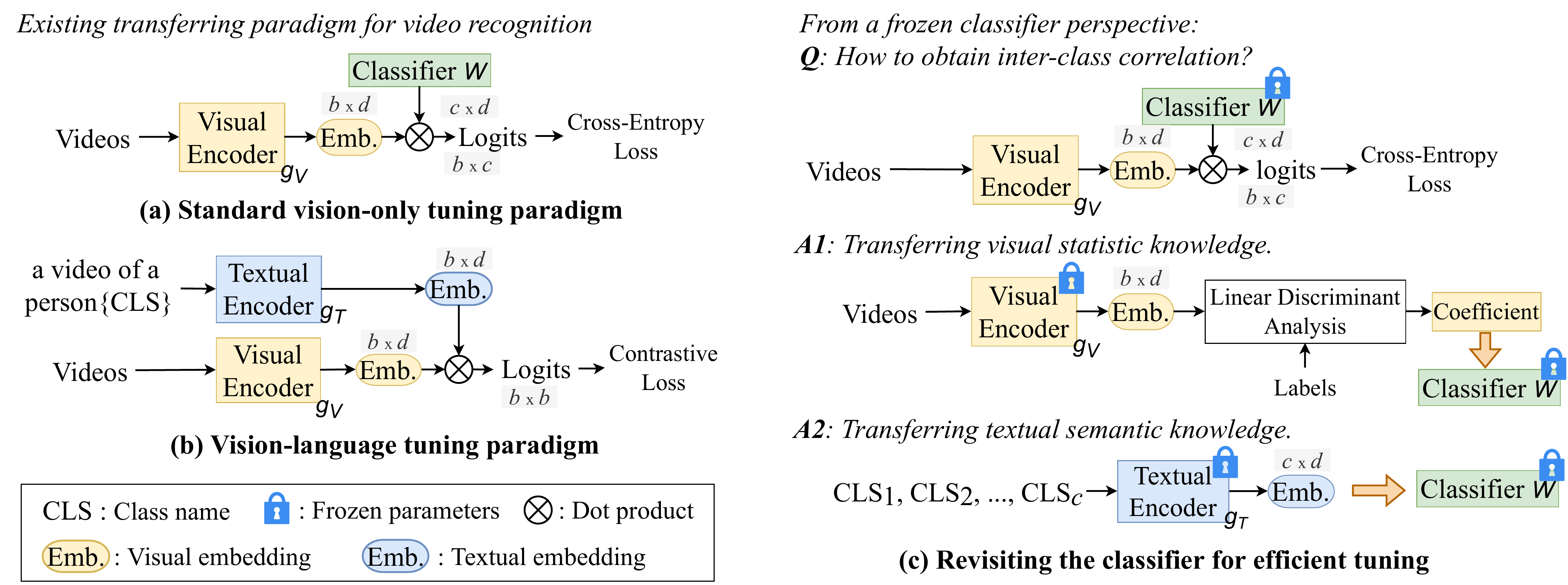}
\end{center}
\caption{Illustration of transferring vision-language pre-trained models for video recognition. (a) The widely-used standard vision-only tuning paradigm with cross-entropy loss. (b) The vision-language tuning paradigm with contrastive loss. (c) Revisiting the role of the classifier to transfer knowledge from vision-language pre-trained models (\eg, CLIP).
}
\label{fig:approach}
\end{figure*}

Next, we examine the final projection head of a vanilla video recognition framework. We conduct the visual-only fine-tuning progress with the visual encoder that is also released by CLIP~\cite{CLIP}. The detailed configurations are provided in \Cref{ab:k400}. The projection head is a matrix of $d \times c$ to compute the pre-softmax values (or logits) from the $d$-dimensional feature vectors for the $c$ classes. Non-rigorously, we can consider the $d$-dimensional row vectors as the embeddings of the class labels, allowing us to explore the inter-class correlation between these learned ``embeddings'', as shown on the right side of Figure~\ref{fig:intro_heatmap}. 
Interestingly, these learned ``embeddings'' also reveal certain correlations after the training, despite being initialized randomly and optimized without knowing any textual information~\footnote{That is, optimized with cross-entropy loss with one-hot labels}.

Therefore, we suppose that the semantic information contained in the samples does correlate with inter-classes. Following this motivation, we replace the projection matrix with several variants: i) The projection matrix whose row vectors are randomly sampled (trivial correlation); ii) The projection matrix whose row vectors are orthogonal to each other (non-correlated); iii) The projection matrix that is initialized using the visual statistic knowledge to provide maximized the correlation between labels (see \Cref{sec:ours}); iv) The projection matrix with fixed embedded textual vectors provides the ``proper'' correlation. In the empirical studies, we find that textual knowledge significantly improves the transferability of pre-trained models, regarding both the classification accuracy and the convergence speed. 
Our main contributions are summarized as follows:
\begin{itemize}
    \item We build a new recognition paradigm to improve the transferability using visual knowledge and textual knowledge from the well-pre-trained vision-language model.
    \item We conduct extensive experiments on popular video datasets (\ie, Kinetics-400 \& 600, UCF-101, HMDB-51 and ActivityNet) to demonstrate the transferability of our solution in many types of transfer learning, \ie, zero-shot / few-shot / general video recognition. 
    Our approach democratizes the training on video datasets and achieves state-of-the-art performance on various video recognition settings, \eg, 87.8\% top-1 accuracy on Kinetics-400, and outperforms previous methods by 20$\sim$50\% absolute top-1 accuracy under zero-shot, few-shot settings.

\end{itemize}

\section{Methodology}

\paragraph{Denotations.}
In the paper, we use bold letters to denote \texttt{Vector}, and capital italic letters to denote \texttt{Tensor} or \texttt{Matrix}, \eg, we employ $\mathbf{z} \in \mathbb{R}^d$ to denote the feature vector extracted from a pre-trained model of dimension $d$, we employ $W \in \mathbb{R}^{d\times c}$ to denote the projection matrix for the $c-$class linear classifier. 
Without ambiguity, we also use capital italic letters to denote the modality in subscripts, especially we employ $V$ and $T$ to denote the \textit{Visual} modality and \textit{Textual} modality, respectively. 
We further employ lowercase italic letters to denote functions or neural networks. For instance, we employ $g_V(\cdot, \Theta_V)$ and $g_T(\cdot, \Theta_T)$ to denote the visual and textual encoder, respectively. 
Besides, we employ calligraphic letters, \eg, $\mathcal{D}$, to denote sets of elements.

\subsection{Revisiting of Previous Tuning Paradigms}

\paragraph{Standard Vision Transferring Paradigm.} As shown in Figure~\ref{fig:approach}(a), we start with the most ordinary scenario, where a visual encoder model $g_V$ is optimized using a large-scale dataset $\mathcal{D}$ that contains visual samples with or without ground-truth labels. On our labeled downstream dataset $\mathcal{\tilde{D}} = \{ (\bm{x}_1, \bm{y}_1), (\bm{x}_2, \bm{y}_2), \ldots \}$, our empirical learning target can be written as
\begin{equation}
    g^*_V, W^* = \underset{\Theta_V,W}{\mathrm{arg min}}~
                {\mathbb{E}}_{{\bm{x},\bm{y} \sim \mathcal{\tilde{D}}}}\big[ H(\bm{y} | \sigma(W \cdot g_V(\bm{x}))) \big], \label{eq:Paradigm}
\end{equation}
where $H(\hat{p}|p)$ stands for the \texttt{CrossEntropy} between the predicted distribution $p$ and the ground-truth distribution $\hat{p}$, $\sigma$ denotes the \texttt{softmax} operation, $W \in \mathbb{R}^{c\times d}$ denotes the linear projection matrix for classification. The formulation in Eq.~\ref{eq:Paradigm} is a standard visual feature transferring paradigm, where the visual encoder $g_V$ and the projection matrix (classifier) $W$ are learned simultaneously.

\paragraph{Vision-Language Learning Paradigm.}
As shown in \Cref{fig:approach}(b), we then review the contrastive learning paradigm of the vision-language models. This paradigm has been widely used for vision-language pre-training \ie, CLIP~\cite{CLIP}, and also been extend to video-text fine-tuning, \ie, ActionCLIP~\cite{wang2021actionclip}, CLIP4Clip \cite{luo2022clip4clip}.
Given a weakly related vision-language pair (\eg, image-text, video-text) dataset $\mathcal{D} = \{ (\bm{x}_{V,1}, \bm{x}_{T,1}), (\bm{x}_{V,2}, \bm{x}_{T,2}) ...\}$. With slight abuse of the notations, we employ the $\bm{x}_V,\bm{x}_T$ to denote a mini-batch of size $b$, then we minimize the following target,
\begin{equation}\label{eq:infonce}
    g^*_V, g^*_T = \underset{\Theta_V,\Theta_T}{\mathrm{arg min}}~
                {\mathbb{E}}_{{\bm{x}_V,\bm{x}_T \sim \mathcal{\tilde{D}}}}\big[H( \mathcal{Q} | \sigma(g_V(\bm{x}_V)^{\textrm{T}}\cdot g_T(\bm{x}_T))) \big],
\end{equation}
where $\mathcal{Q}$ is the set that contains $b$ one-hot labels of size $c$, with their $1, 2, \ldots, b$-th element being $1$ ($b<c$, denoting the positive vision-language pairs. Here we clarify that, the definition in Eq.~\ref{eq:infonce} is not the rigorous form of the Noise-Contrastive Estimation (NCE) loss proposed in~\cite{infonce}. Instead, we employ the cross entropy version implementation in~\cite{CLIP,chen2021empirical}. This implementation depicts a connection between the standard feature transferring paradigm and ours. In which the $g_T(\bm{x}_T)$ can be considered as the projection matrix that map the visual feature $g_V(\bm{x}_V)$ to the given label set $\mathcal{Q}$.

\subsection{Our Proposed Paradigm}\label{sec:ours}
As discussed in \Cref{sec:intro}, we replace the learnable randomly initialized linear projection matrix $W$ with pre-defined matrix $\tilde{W}$. Similarly, the training target can be written as 
\begin{equation}
    g^*_V = \underset{\Theta_V}{\mathrm{arg min}}~
                {\mathbb{E}}_{{\bm{x},\bm{y} \sim \mathcal{\tilde{D}}}}\big[ H(\bm{y} | \sigma(\tilde{W}\cdot g_V(\bm{x}))) \big].
\end{equation}
Note that $\tilde{W}$ is not in the optimization targets, since we freeze it from updating during the fine-tuning of the downstream tasks. We do this for two reasons: Firstly, it could preserve the textual knowledge from being disturbed by the randomness brought by the mini-batch. For instance, when some classes are missing, their embedded feature vector might be broken by the other classes; Secondly, we want to provide a fair comparison between different initialization of $\tilde{W}$.
Now we consider how to initialize $\tilde{W}$. To examine how the correlation between the semantic information contained in the samples helps, we investigate the following four types of initialization, which represent different degrees of inter-class correlation. 

\textbf{Randomized Matrix.}
For the most simple randomized matrix case, we set each row of the $\tilde{W}$ with a random Gaussian vector of zero mean and standard deviation, that is
\begin{equation}
\tilde{W} \sim \mathcal{N}(\bm{0}, I_d),
\end{equation}
where $I_d$ denotes the identity matrix of dimension $d\times d$. Arithmetically, a trivial ``correlation'' would appear between the row of the $\tilde{W}$, since the sampling size is significantly small to be biased. Evidently, the trivial ``correlation'' cannot indicate the real correspondence between the classes due to its stochasticity. Therefore we expect the model to have inferior performance since it needs to avoid these incorrect correlations when learning the visual feature representation.

\textbf{Randomized Orthogonal Matrix.}
We follow the approach of the randomized matrix. We then remove the correlation by ensuring the row vectors are orthogonal. This is achieved by QR decomposition. Concretely, since $d>c$, we first generate a random matrix of size $d \times d$ and select the first $c$ rows as our projection matrix. Formally, we have,
\begin{equation}
\begin{aligned}
\tilde{W}_j \sim \mathrm{QR}(U)_j, j = 1, 2, \ldots, c, \\ 
U_i\sim \mathcal{N}(\bm{0}, I_d), i = 1, 2, \ldots, d,
\end{aligned}
\end{equation}
where $U$ is the intermediate randomized matrix, $\mathrm{QR}(U)$ is the row orthogonal matrix obtained through the QR decomposition.
Similar to the randomized matrix, we also expect this initialization to have inferior performance. Given the fact that the one-hot label vectors are also orthogonal to each other, it will not be helpful to project the visual feature vectors with an orthogonal matrix, which increases the difficulty of learning meaningful visual features.

\textbf{Linear Discriminant Projection.}\label{sec:lda}
We consider another way of initializing the projection matrix. We employ the multi-class Fisher's linear discriminant analysis (LDA) to learn a linear classifier, then employ the weight matrix of the classifier as our initialization of the projection matrix. 
Specifically, we use the pre-trained visual encoder to extract visual embeddings of samples in the train split, then perform LDA on the pre-extracted visual embeddings of the training set to generate the LDA coefficient. Finally, we use the LDA coefficient to initialize $\tilde{W}$ and freeze it for fine-tuning the visual encoder on the dataset.
We compute the LDA projection following previous work~\cite{lda_coef}.
Intuitively, the LDA simultaneously maximizes the inter-class covariance and minimizes intra-class covariance.
We, therefore, term this as the maximal correlation initialization using the visual statistic knowledge. As an essential classifier, this type of initialization delivers reasonable performance, but it is largely dependent on the data employed to compute the projection matrix. When the data is limited, the estimated correlation will be biased. 
On the other hand, in our proposed paradigm, the pre-trained textual encoder provides unbiased correlations for fine-tuning.

\textbf{Textual Embedding Vectors.} 
We finally describe the paradigm to transfer textual semantic knowledge from a pre-trained textual encoder. Briefly, the projection weight $\tilde{W}$ is composed of the embedded textual feature vectors of the labels. Given a set of tokenized class labels $\mathcal{L} = \{\bm{l}_1, \bm{l}_2, \ldots, \bm{l}_c \}$, we have 
\begin{equation}
\tilde{W}_i \sim g_T(\bm{l}_i), i = 1, 2, \ldots, c,
\end{equation}
where $\tilde{W}_i$ the $i$-th row vector in matrix $\tilde{W}$. And $\tilde{W}_i$ is initialized using the textual encoder output of the textual label of the $i$-th class. 
In the experimental analysis, we investigate two types of textual feature encoders: i) The encoder that is trained with a visual encoder in the contrastive style, \ie, CLIP; ii) The encoder that is trained solely using only textual samples on tasks such as masked language modeling, \ie, DistilBERT~\cite{sanh2019distilbert}.

 

\section{Related Works}
\paragraph{Visual Recognition.} 
Convolutional networks have long been the standard for backbone architectures in image recognition~\cite{alexnet,resnet,vgg,bn} and video recognition~\cite{i3d,p3d,s3d,r2+1d}.
Inspired by the Transformer~\cite{vaswani2017attention} scaling successes in Natural Language Processing, Vision Transformer (ViT)~\cite{ViT} applies a standard Transformer directly to images, which delivers impressive performance on image recognition.
Since then, ViT~\cite{ViT} has led a new trend in image recognition backbone architectures, shifting from CNNs to Transformers.
To improve performance, follow-up studies, \eg, DeiT~\cite{DeiT}, Swin~\cite{liu2021swin}, have been developed.
Also, many works has begun to adopt transformers in video recognition, such as TimeSFormer~\cite{timesformer}, ViViT~\cite{arnab2021vivit}, VideoSwin~\cite{videoswin}, and MViT~\cite{mvit}.

\paragraph{Image-Language Pre-training.}\label{clip}
Recently, CLIP~\cite{CLIP} provides good practice in learning the coordinated vision-language pretraining models using the image-text InfoNCE contrastive loss~\cite{infonce}. 
Based on CLIP, several variants~\cite{ALIGN, li2022blip,yuan2021florence,yu2022coca} have been proposed by combining more types of learning tasks such as image-text matching and masked image/language modeling. These contrastively learned models have two deserved properties for downstream tasks: the abundant visual feature representations and the aligned textual feature representations. Yet another study~\cite{yang2022unified} merged the downstream classification task into the pretraining progress, which demonstrates a decent improvement of accuracy over the standard cross-entropy loss.

\paragraph{Transferring CLIP Models for Video-Text Learning.}
Recently, many video-text retrieval methods~\cite{wang2021t2vlad,zhao2022centerclip,luo2022clip4clip,wu2022cap4video} have benefited from vision-language pre-training as well.
Moreover, several recent works~\cite{wang2021actionclip,ju2021prompting,bike} extend the CLIP~\cite{CLIP} to train a downstream video-text matching model with contrastive loss, then perform video recognition using the similarity between learned video and text embeddings during inference.
Instead of these contrastive-based methods, we investigate the correlations of the linear classifier for efficient feature transferring in the standard visual recognition paradigm. 
Then we directly transfers visual and textual knowledge for video recognition. 
In comparison to contrastive-based methods, we demonstrate the superiority of our method in efficient training in \Cref{tab:info}. 
We hope that the simple and effective paradigm can serve as a new baseline for future work.
%

\section{Experiments: Video Recognition}
\subsection{Setups}
To evaluate our method for video recognition, we conduct experiments on five popular datasets, \ie, Kinetics-400~\cite{kay2017kinetics}, Kinetics-600~\cite{k600},  UCF-101~\cite{ucf101}, HMDB-51~\cite{hmdb} and ActivityNet-v1.3~\cite{caba2015activitynet}. 
\emph{See Supplementary for statistics of these datasets}.

\begin{table*}[ht]
    \centering
    \scalebox{0.95}{
  \begin{tabular}{lcccccc}
\toprule
  Method & Input   & Pre-train & Top-1 & Top-5 & FLOPs\x Views  & Param  \\
  \midrule
  NL I3D-101~\cite{nonlocal} & 128\x224$^2$  & IN-1K & 77.7 & 93.3 & 359\x10\x3 & 61.8\\
  MVFNet$_{En}$~\cite{wu2020MVFNet}  & 24\x224$^2$  & IN-1K & 79.1 & 93.8 & 188\x10\x3 & - \\
  SlowFast NL101~\cite{slowfast} & 16\x224$^2$  & Scratch & 79.8 & 93.9 & 234\x10\x3 & 59.9 \\
  X3D-XXL~\cite{feichtenhofer2020x3d} & 16\x440$^2$  & Scratch & 80.4 & 94.6 & 144\x10\x3 &20.3 \\
  \hline
  
 \rowcolor{gray!20}
 \multicolumn{7}{l}{\emph{Methods with large-scale pre-training}} \\
  TimeSformer-L~\cite{timesformer} & 96\x224$^2$  & IN-21K & 80.7 & 94.7 & 2380\x1\x3 & 121.4\\
  ViViT-L/16\x2~\cite{arnab2021vivit} & 32\x320$^2$  & IN-21K & 81.3 & 94.7 & 3992\x4\x3 & 310.8\\  
  VideoSwin-L~\cite{videoswin} & 32\x384$^2$  & IN-21K & 84.9 & 96.7 & 2107\x10\x5 & 200.0 \\  
  ip-CSN-152~\cite{CSN} & 32\x224$^2$  & IG-65M  & 82.5 & 95.3 & 109\x10\x3 & 32.8 \\
  ViViT-L/16\x2~\cite{arnab2021vivit} & 32\x320$^2$ & JFT-300M & 83.5 & 95.5 &   3992\x4\x3 & 310.8\\
  TokLearner-L/10~\cite{ryoo2021tokenlearner} & 32\x224$^2$  & JFT-300M & 85.4 & 96.3 & 4076\x4\x3 & 450 \\
  MTV-H~\cite{MTV} & 32\x224$^2$  & JFT-300M & 85.8 & 96.6 & 3706\x4\x3 & - \\
  CoVeR~\cite{cover} & 16\x448$^2$  & JFT-300M & 86.3 & - & -\x1\x3 & - \\
  Florence~\cite{yuan2021florence} & 32\x384$^2$  & FLD-900M & 86.5 & 97.3 & -\x4\x3 & 647 \\ 
  CoVeR~\cite{cover} & 16\x448$^2$  & JFT-3B & 87.2 & - & -\x1\x3 & - \\ 
  VideoPrompt~\cite{ju2021prompting} & 16\x224$^2$  & WIT-400M & 76.9 & 93.5 & - & - \\ 
  ActionCLIP~\cite{wang2021actionclip} & 32\x224$^2$  & WIT-400M & 83.8 & 96.2 & 563\x10\x3 & 141.7 \\ 
  \midrule
  Ours ViT-L/14  & 32\x224$^2$  & WIT-400M & 87.1 & 97.4 & 1662\x4\x3 & 230.7 \\ 
  Ours ViT-L/14  & 32\x336$^2$  & WIT-400M & \baseline{\textbf{87.8}} & \baseline{\textbf{97.6}} & 3829\x1\x3 & 230.7 \\ 
\bottomrule
  \end{tabular}  } 
 \caption{Comparisons with SOTAs on Kinetics-400. ``Views'' indicates \# temporal clip $\times$ \# spatial crop. The magnitudes are Giga ($10^{9}$) and Mega ($10^{6}$) for FLOPs and Param. “IN” denotes ImageNet. }
    \label{tab:k400_sota}
\end{table*}

\begin{table}[t]
\centering
\scalebox{0.95}{
\begin{tabular}{lcc} \toprule
  Method   & Top-1 & mAP \\ \midrule
   ListenToLook~\cite{gao2020listen}  & - & 89.9 \\
   MARL~\cite{wu2019multi} & 85.7 & 90.1 \\ 
   DSANet~\cite{dsanet} & - & 90.5 \\ 
   TSQNet~\cite{tsqnet} & 88.7 & 93.7 \\ 
   NSNet~\cite{nsnet} & 90.2 & 94.3 \\        
   \midrule
   Ours ViT-L & \textbf{92.9} & \textbf{96.5} \\ 
   Ours ViT-L (336$\uparrow$) & \baseline{\textbf{93.3}} & \baseline{\textbf{96.9}} \\ 
   \bottomrule
\end{tabular} }
 \caption{Comparisons with SOTAs on ActivityNet.}
\label{tab:anet_sota}
\end{table} 
    
\begin{table}[t]
\centering
\scalebox{0.94}{
\setlength{\tabcolsep}{2.0pt}
\begin{tabular}{lccccc} \toprule
  Method   & shot & HMDB & UCF & ANet & K400 \\ \midrule
   VideoSwin~\cite{videoswin} & 2 & 20.9  & 53.3  & - & - \\
   VideoPrompt~\cite{ju2021prompting}  & 5  & 56.6  & 79.5  & - & 58.5 \\
   X-Florence~\cite{XCLIP}  & 2  & 51.6  & 84.0  & - & - \\ \midrule
   \multirow{4}{*}{Ours ViT-L}
     & 0  & 53.8  & 71.9  & 75.6 & 61.0 \\
      & 1  & \baseline{\textbf{72.7}}  & \baseline{\textbf{96.4}}  & \baseline{\textbf{89.0}} & \baseline{\textbf{75.8}} \\
    & 2  & \baseline{\textbf{73.5}}  & \baseline{\textbf{96.6}}  &  \baseline{\textbf{90.3}} & \baseline{\textbf{78.2}} \\
      & All  & 80.1   & 96.9   & 91.1  & 84.7 \\
    \bottomrule
\end{tabular} }
\caption{Comparisons with SOTAs on few-shot recognition.}
\label{table:video_few}
\end{table}

\paragraph{Training \& Inference.}\label{training}
The video recognition task takes a video as input, and then fed it into a learned encoder to estimate the action category of the video. 
Given a video, we first uniformly sample $T$ (\eg, 8, 16, 32) frames over the entire video. 
Then we utilize ResNet~\cite{resnet} or ViT~\cite{ViT} as the video encoders.
The classifier in our paradigm is intialized from the textual embedding of the class names and then frozen (fixed), leaving only the parameters in the video encoder to be learned. 
To trade off accuracy and speed, we consider two inference strategies: 
(1) \emph{Single View}: We use only 1 clip per video and the center crop for efficient evaluation, (\eg, as in \Cref{ab:k400}).
(2) \emph{Multiple Views}: This is a widely used setting in previous works~\cite{slowfast,i3d} to sample multiple clips per video with several spatial crops in order to get higher accuracy. For comparison with SOTAs, we use four clips with three crops (``4\x3 Views'') in Table~\ref{tab:k400_sota}.
\emph{See Supplementary for training hyperparameters}.

\begin{table*}[t]
    \centering
    	\begin{tabular}{lllcc}
    	\toprule
    	Method  & UCF$^*$ / UCF & HMDB$^*$ / HMDB & ANet$^*$/ ANet & Kinetics-600 \\ \midrule
    	GA~\cite{GA} & 17.3$\pm$1.1 / - & 19.3$\pm$2.1 / - & -  & -  \\
    	TS-GCN~\cite{TS-GCN} & 34.2$\pm$3.1 / - & 23.2$\pm$3.0 / - & - & - \\
        E2E~\cite{E2E} & 44.1 / 35.3 & 29.8 / 24.8 & 26.6 / 20.0 & - \\
        DASZL~\cite{DASZL} & 48.9$\pm$5.8 / - & - / - & - & -\\
        ER~\cite{ER} & 51.8$\pm$2.9 / - & 35.3$\pm$4.6 / - & - & 42.1$\pm$1.4 \\
        ResT~\cite{ResT} & 58.7$\pm$3.3 / 46.7 & 41.1$\pm$3.7 / 34.4 & 32.5 / 26.3 & - \\
        \midrule
        Ours & \baseline{\textbf{85.8$\pm$3.3 / 79.6}}  & \baseline{\textbf{58.1$\pm$5.7 / 49.8}}  & \baseline{\textbf{84.6$\pm$1.4 / 77.4}} & \baseline{\textbf{68.9$\pm$1.0}} \\
    	\bottomrule
    	\end{tabular} 
 \caption{Comparisons with SOTAs on zero-shot video recognition. We directly evaluate our method without any additional training on cross-dataset video recognition. ANet is in short for ActivityNet. $^*$ means half classes evaluation.}
    \label{tab:sota_zero}        
\end{table*}

\subsection{Main Results}

\paragraph{Comparison to State-of-the-Arts.} 
In Table~\ref{tab:k400_sota}, on the challenging \textbf{Kinetics-400} dataset, we compare to state-of-the-arts that are pre-trained on large-scale datasets such as ImageNet-21K~\cite{deng2009imagenet}, IG-65M~\cite{ig65m}, JFT-300M~\cite{JFT300M}, FLD-900M~\cite{yuan2021florence} and JFT-3B~\cite{JFT3B}. 
Up to now, none of the three largest datasets (\ie, JFT-300M, FLD-900M, JFT-3B) is open-sourced and also does not provide pre-trained models.
Thus, we use the CLIP~\cite{CLIP} checkpoints, which are publicly available\footnote{https://github.com/openai/CLIP/blob/main/clip/clip.py} and have been trained on 400 million web image-text pairs (namely WIT-400M).
We can observe that our model outperforms all JFT-pretrained methods in terms of Top-1 and Top-5 accuracy. 
We achieve an accuracy of 87.8\% , which improves even further by 1.3\% over Florence~\cite{yuan2021florence}, although their model and data scale are both 2$\times$ larger than ours.
Besides, our model is even better than CoVeR~\cite{cover}, and their data scale is 7.5$\times$ larger.

To verify the generalization ability of our method, we further evaluate the performance of our method on the well-known untrimmed video benchmark, \textbf{ActivityNet-v1.3}. We finetuned the Kinetics-400 pre-trained models with 16 frames on the Activitynet-v1.3 dataset and report the  top-1 accuracy and mean average precision (mAP) following the official evaluation metrics. As shown in Table~\ref{tab:anet_sota}, our method outperforms recent SOTAs with a clear margin. To the best of our knowledge, our method achieves the best performance (96.9\%) on ActivityNet.
We also evaluate our method on the \textbf{UCF-101} and \textbf{HMDB-51} datasets to demonstrate its capacity to generalize to smaller data. We achieve the mean class accuracy of 98.2\% on UCF and 81.3\% on HMDB, respectively. 
\emph{Please see supplementary for more comparisons on UCF-101 and HMDB-51.}

\paragraph{Few-Shot Video Recognition.}\label{exp:video_fewshot}
Video recognition using only a few samples is known as few-shot video recognition.
We study a more challenging $K$-shot $C$-way situation instead of the conventional 
5-shot 5-way configuration. 
We scale the task up to categorize \textbf{all} categories in the dataset with just $K$ samples per category for training. 
The lower and upper bound of this situation are denoted by the term ``Zero-shot'' and ``All-shot'' respectively.
Table~\ref{table:video_few} reports the Top-1 accuracy for the four datasets.
In this extreme scenario of few data, we use CLIP-pretrained ViT-L/14 with 8 frames and TAP for few-shot video recognition.
In these extremely data-poor situations (\eg, even with just one shot), we can see that our method offers amazing transferability to diverse domain data.
Our approach, in contrast, demonstrates robustness by outperforming SOTAs by a large margin. For instance, when comparing accuracy on HMDB-51 with 2-shot, our method outperforms Swin, X-Florence by \textbf{+52.6\%} and \textbf{+21.9\%} respectively.
\emph{See Supplementary for training details.}

\paragraph{Zero-Shot Video Recognition.}\label{exp:zero_video}
Furthermore, we conduct experiments in the open-set setting. We use our Kinetics-400 pre-trained models (\ie, ViT-L with 8 frames) to perform the zero-shot evaluation on four other video datasets.
On UCF-101, HMDB-51 and ActivityNet, there are two major evaluation protocols following \cite{E2E}: half classes evaluation and full classes evaluation.
\emph{Please see Supplementary for the details of two evaluation protocols and the Kinetics-600 evaluation.}
We present comprehensive comparisons on four datasets in Table~\ref{tab:sota_zero}, our method shows a strong cross-dataset generalization ability.
Our method shows a large improvement upon previous zero-shot recognition methods (\textbf{+27.1\%} on UCF-101, \textbf{+17.0\%} on HMDB-51, \textbf{+52.1\%} on ActivityNet, \textbf{+26.8\%} on Kinetics-600).

\subsection{Ablations on Kinetics}\label{ab:k400}
In this section, we conduct extensive ablation experiments on the Kinetics-400 dataset.
Unless specified otherwise, we use ViT-B/16 with 8 frames as the video backbone and a single view for testing. The default settings are marked in \colorbox{baselinecolor}{gray}.
\emph{See Supplementary for more ablations.}

\paragraph{Different Initializations to the Offline Classifier.}
We set different initializations described in \Cref{sec:ours} to the offline classifier $W \in \mathbb{R}^{d\times c}$ and then train our visual encoder on Kinetics-400. Table~\ref{table:init} lists their comparisons. 
We show that feeding the offline classifier a random $d$-by-$c$ matrix with a normal distribution reduces performance significantly.
Then we assign the orthogonal matrix to the classifier,
and see that removing the inter-class correlation of the classifier will result in inferior performance.
Furthermore, we term the linear discriminate projection as the maximal correlation initialization. 
To do so, we first sample 60 videos from each class in the training set and utilize the pre-trained visual encoder to extract visual embeddings from these 24,000 videos. 
Finally, we learn the linear classifier by performing linear discriminant analysis on these visual embeddings and their labels. 
We can see the LDA projection achieves a strong baseline.

Finally, we study the textual embeddings from different textual encoders.
We choose DistilBERT~\cite{sanh2019distilbert} and CLIP~\cite{CLIP} as the textual encoder to pre-extract the text embeddings of $c$ categories. We observe that DistilBERT performs the same performance as CLIP's textual encoder. This may be because both DistillBERT and CLIP are pre-trained with large-scale data, so they both have strong language modeling capabilities and can generate good semantic targets. Although the good semantic targets generated by DistillBERT are not aligned with the visual features of CLIP, it is easy to fit them with trainable visual encoders. We also observe that the loss of DistillBERT will be higher than CLIP in the early stage, but it will quickly decrease to the same level.
\emph{More visualizations of these classifiers are in Supplementary.}

\begin{table}[t]
    \centering
      \begin{tabular}{lc}
      \toprule
      Offline classifier from & Top 1 \\
      \midrule
      Random normal matrix & 59.3 \\
      Random orthogonal matrix & 59.4  \\
      Linear discriminant projection & 80.8  \\
      DistilBERT & 81.4  \\
      Textual encoder of CLIP & \baseline{\textbf{81.5}} \\
      \bottomrule
       \end{tabular}
       \caption{Exploration of different frozen classifiers.}
      \label{table:init}  
\end{table}

\begin{table}[t]
    \centering
  \begin{tabular}{cccc}
  \toprule
    & Zero-shot & 2-shot & Full-shot  \\ \midrule
  \emph{Vision-Only} & 0.2  & 21.6 & 75.3\\ 
  \emph{Vision-Text} & \textbf{54.2} &  \textbf{65.3} & \textbf{80.1}   \\ \bottomrule
   \end{tabular}
   \caption{Comparisons with vision-only framework.}
  \label{table:vs_vision}
\end{table}

\paragraph{Comparison with Vision-Only Tuning Paradigm.} 
As a comparison with our method, we train the unimodality video model, which consists of the same visual encoder and a learnable classifier with random initialization.
To produce video embedding, we just apply temporal average pooling (TAP) to frame embeddings.
As shown in Table~\ref{table:vs_vision}, our \emph{Vision-Text} method leads to obvious improvement with the same training recipe, especially in the data-poor situation.

\paragraph{Temporal Modeling.}
Here we explore more temporal modelings for ViT and ResNet:
(1) \textbf{TAP}: Temporal average pooling is the most straightforward temporal modeling.
(2) \textbf{T1D}: The channel-wise temporal 1D convolutions, is a common strategy~\cite{wu2020MVFNet,tdn,teinet}, to perform efficient temporal interaction in the latter stages (\ie, res$_{4-5}$) of ResNet.
(3) \textbf{T-Trans}: The embeddings of frames are fed to a multi-layer (\eg, 6-layer) temporal transformer encoder. 
(4) \textbf{TokenT1D}: We use T1D to model temporal relations for [class] token features that are aggregated from local features via attention in the vision transformer. We perform the TokenT1D in multiple positions of a vision transformer.
Results are shown in Table~\ref{table:T-Model}. 
On both backbones, TAP provides simple baselines and T-Trans exhibits the best top-1 accuracy. Both of them maintain the original frame-level representations and then perform temporal modeling.
An interesting thing we observed is that T1D does not seem to work in this scenario. The reason lies in that T1D may have the potential to break the learned strong representations provided by CLIP.
TokenT1D is another internal-backbone temporal modeling, and it does not yield a performance drop, and even slightly improves the TAP baseline.
We believe this is because TokenT1D is only imposed on the global [class] token instead of patch tokens, resulting in minimal modifications on pre-trained features.

\begin{table}[t]
    \centering
    \begin{tabular}{cccc} 
    \toprule
     Backbone & Modeling & Top-1 & Top-5 \\ \midrule
     \multirow{3}*{ResNet-50} & TAP & 71.2 & 90.4 \\
      & T1D & 67.2 & 88.5 \\
      & T-Trans & \textbf{74.3} & \textbf{91.7} \\ \midrule
     \multirow{3}*{VIT-B/16} & TAP & 80.1 & 95.0 \\
      & TokenT1D & 80.4 & 95.0 \\
      & T-Trans & \baseline{\textbf{81.5}} & \baseline{\textbf{95.5}} \\\bottomrule 
    \end{tabular} 
    \caption{Temporal modeling for video encoders.}
    \label{table:T-Model} 
\end{table}

\begin{table}[b]
    \centering
  \begin{tabular}{ccccc}
  \toprule
  Paradigm  & \makecell[c]{Batch \\ Gather} & \makecell[c]{Textual \\ Encoder} & Top-1   & V100-days \\
  \midrule
  \multirow{4}*{\makecell[c]{Contrastive- \\ Based}}  & \cmark & online & 81.2 & 6.7 (10$^*$) \\
   & \cmark & offline & 80.7  & 6.6 \\
   & \xmark & online & 77.8  & 3.5 \\
   & \xmark & offline & 76.1  & 3.3 \\ \midrule
  \multirow{1}*{Ours}  & \xmark & offline & \baseline{\textbf{81.5}}  & \textbf{3.3} \\
  \bottomrule
  \end{tabular}
    \caption{Ours \emph{vs.} Contrastive-based paradigm with ViT-B/16 on Kinetics-400. The number of V100 days is the number of V100 GPU used for training multiplied by the training time in days. $*$ indicates the official result~\cite{wang2021actionclip} via ``Data-parallel training'' on 3090 GPUs. For efficient training and fair comparison, we implement all experiments with ``Distributed Data-parallel training'' in this Table.}  
  \label{tab:info}
\end{table}

\begin{table}[t]
    \centering
    \begin{tabular}{ccc} 
    \toprule
     Views  & Top-1 & GFLOPs \\ \midrule
      Single$\rightarrow$Multiple  & 81.5$\rightarrow$\textbf{82.9} & \textbf{90.3}$\rightarrow$90.3\x12  \\
    \bottomrule 
    \end{tabular} 
    \caption{Two classic evaluation protocols.}
    \label{table:views} 
\end{table}


\begin{table}[t]
    \centering
    \footnotesize
    \begin{tabular}{lccccc} 
    \toprule
      Method    & Top-1 & FLOPs & Params & Throughput \\ \midrule
ViViT-L/16-320	& 81.3  &	3992G &	310.8M	& 4.2 vid/s$^*$ \\ 
Ours ViT-B/32 	& 78.5  &	23.7G &	71.6M	& 322.5 vid/s \\
Ours ViT-B/16 	& 81.5  &	90.3G &	69.9M	& 126.5 vid/s \\
Ours ViT-L/14 	& 85.4  &	415.4G &	230.4M &	35.5 vid/s \\ \bottomrule
    \end{tabular}
    \caption{Analysis on throughput. ``vid/s" represents the average number of videos per second. The larger ``vid/s" represents higher efficiency. $^*$ is the official result with TPU-v3. }
    \label{table:runtime} 
\end{table}

\paragraph{Ours \emph{v.s.} Contrastive-Based Paradigm.}
We make a comparison with the Contrastive-based tuning method \ie, ActionClip~\cite{wang2021actionclip} mentioned in \Cref{clip}. This paradigm treats the recognition task as a video-text matching problem with contrastive loss, thus requiring a batch gathering to collect embeddings of all batches across all GPUs and calculate cosine similarity for a given batch across all other batches. 
In Table~\ref{tab:info}, we compare it with the Contrastive-based paradigm and
observe that it does not work well without batch gathering. This is due to contrastive learning favors a large batch size (\emph{e.g.,} CLIP used 256 GPUs with a batch size of 128 per GPU to maintain a large 32768\x32768 similarity matrix).
Besides, involving batch gather will multiply the training time.
Also, in this case, the pre-trained textual encoder still needs to be updated, which requires larger GPU memory.
However, our paradigm employs pre-extracted text embeddings as our classifier, so the only learned part is the visual encoder.
Results show that our method achieves the best accuracy-cost trade-off. Specifically, our method achieves the performance of 81.5\% with ViT-B/16, which takes only 10 hours to run the training using 8 GPUs (\textbf{2$\times$ faster} than the matching counterpart).
\emph{See Supplementary for details about the batch gathering.}

\paragraph{More Instantiations.} 
Table~\ref{table:runtime} presents the results of our method using different visual encoders, indicating that deeper backbones can achieve better performance. Table~\ref{table:views} presents the results of our method under two evaluation protocols mentioned in \Cref{training}, where the multi-view evaluation protocol results in additional improvements.


\paragraph{Analysis on Efficiency.}
In Table~\ref{table:runtime}, we present the computational cost and efficiency of our models. We follow the common inference settings by using a single NVIDIA A100 GPU to measure the throughput. We use a batch size of 16 to measure the throughput. 
Our models achieve the \textbf{29$\times$ faster} throughput and \textbf{44$\times$ fewer} FLOPs compared with the previous transformer-based method ViViT~\cite{arnab2021vivit} under the same accuracy.




\section{Limitation and Conclusion}
\textbf{Limitation}:
The performance of the proposed paradigm is restricted to how the category labels are represented. For instance, in tasks such as human re-identification, the labels are often set as numerical values such as 0, 1, 2, etc. In this case, we cannot transfer any semantic information from the textual encoders, while transferring visual statistic knowledge (\ie, LDA classifier) could be helpful.

\noindent\textbf{Conclusion}:
We present a new paradigm for improving the transferability of visual recognition that is based on the knowledge from the textual encoder of the well-trained vision-language model.
The empirical study shows that our method improves both the performance and the convergence speed of visual classification. 
The proposed approach has superior performance on both general and zero-shot/few-shot recognition and achieves state-of-the-art performance on video recognition tasks, and democratizes transferring on challenging video datasets, \ie, Kinetics-400.

\section*{Acknowledgments}
This paper is supported by the Australian Research Council Grant DP200103223, Australian Medical Research Future Fund MRFAI000085, CRC-P Smart Material Recovery Facility (SMRF) – Curby Soft Plastics, and CRC-P ARIA - Bionic Visual-Spatial Prosthesis for the Blind.

\bibliography{aaai23}

\clearpage
\appendix
\section*{Appendix}
\setcounter{table}{0}
\setcounter{figure}{0}
\renewcommand{\thetable}{A.\arabic{table}}
\renewcommand{\thefigure}{A.\arabic{figure}}

In this appendix, 
\S\ref{supp:imp_details} contains additional \textit{details} for: the training details (\S\ref{sec:train_video}), zero-shot evaluation~(\S\ref{sec:zero}), the statistics of video datasets (\S\ref{sec:video_datasets}), details of several large-scale datasets for pre-training (\S\ref{sec:large-scale}), visual encoder architectures (\S\ref{sec:vis_encoder}), Batch Gather (\S\ref{sec:batch_gather}), LDA (\S\ref{supp:lda}) and data overlaps (\S\ref{sec:overlaps}).
\S\ref{sec:results} contains further \textit{results} for video recognition: comparison with SOTAs on UCF-101 and HMDB-51 (\S\ref{sec:small_sota}), more visualizations of different classifier (\S\ref{sec:vis}) and more ablations (\S\ref{sec:ab}).

\section{Additional Details}\label{supp:imp_details}

\subsection{Training details}\label{sec:train_video}
\textbf{General video recognition}: 
In Table~\ref{tb:imp}, we present our training details for general video recognition. We share the same recipe on all the video datasets, \ie, Kinetics-400, ActivityNet, HMDB-51, UCF-101.

\noindent\textbf{Few-shot video recognition}:
We repeat the samples to keep the same iterations with the above general counterpart. For example, we train the model on Kinetics-400 with $\sim$900 iterations per epoch for general setting, we repeat the sample to maintain the same $\sim$900 iterations per epoch for few-shot setting. In this way, we only the few-shot models with 2 epochs on Kinetics-400, and 10 epochs on other video datasets, \ie, ActivityNet, HMDB-51, UCF-101. Other settings are same with the Table~\ref{tb:imp}.

\noindent\textbf{Zero-shot video recognition}: We use the Kinetics-400 pre-trained models to directly perform cross-dataset zero-shot video recognition \textbf{without any additional training} on other datasets \ie, ActivityNet, HMDB-51, UCF-101 and Kinetics-600.

\begin{table}[h!]
\centering
\begin{tabular}{l|c} \toprule
    Setting  & Value \\ \midrule
    \multicolumn{2}{l}{\cellcolor{mygray}\emph{Training Hyperparameter}} \\
    Batch size & 256 \\
    Vocabulary size & 49408 \\
    Training epochs & 30 \\
    Optimizer & AdamW \\
    Learning rate (base, minimal) & (5e-5, 5e-6), cosine \\
    Weight decay & 0.2 \\
    Linear warm-up epochs & 5 \\
    Adam $\beta_1$,$\beta_2$ & 0.9, 0.98 \\ \midrule
    \multicolumn{2}{l}{\cellcolor{mygray}\emph{Augmentation}} \\
    Resize & RandomSizedCrop \\
    Crop size & 224 (Default) \\
    Random Flip & 0.5 \\
    Random Gray scale & 0.2 \\
    RandAugment & $N=2$, $M=9$ \\
    \bottomrule
\end{tabular}
\caption{Default training details for video recognition}
\label{tb:imp}
\end{table}

\subsection{Evaluation protocols of zero-shot video recognition}\label{sec:zero}
We use our Kinetics-400 pre-trained models to perform zero-shot evaluation on other datasets.
On UCF-101, HMDB-51 and ActivityNet, there are two major evaluation protocols following \cite{E2E}: 
\begin{itemize}
    \item[1.] In order to make our results comparable with previous works, we randomly choose half of the test dataset’s classes, 50 for UCF, 25 for HMDB, and 100 for ActivityNet. Evaluate on the selected subset. Repeat ten times and average the results for each test dataset. We donate this setting as UCF$^*$, HMDB$^*$ and ANet$^*$.
    \item[2.] The second evaluation setting is directly evaluating on the full dataset. This allows us to return more realistic accuracy scores. 
\end{itemize}

On Kinetics-600, we follows~\cite{ER} to choose the 220 new categories outside Kinetics-400 in Kinetics-600 for evaluation. We use the three splits provided by ~\cite{ER}. For each split, we sample 160 categories for evaluation from the 220 categories in Kinetics-600. We report the mean accuracy for three splits.

\subsection{Statistics of video datasets}\label{sec:video_datasets}
\begin{itemize}
\item \textbf{Kinetics-400}~\cite{kay2017kinetics} is a large-scale video dataset, which consists of 240k training videos and 20k validation videos in 400 different human action categories. Each video in the dataset is a 10-second clip of action moment annotated from raw YouTube video.
\item \textbf{UCF-101}~\cite{ucf101} contains 13k videos spanning over 101 human actions.
\item \textbf{HMDB-51}~\cite{hmdb} contains approximately 7k videos belonging to 51 action class categories.
\item \textbf{Kinetics-600} is an extensions of the Kinetics-400 dataset. Kinetics-600 consists of around 480k videos from 600 action categories. The 480K videos are divided into 390k, 30k, 60k for training, validation and test sets, respectively. In this paper, we use its test set for zero-shot evaluation.
\item \textbf{ActivityNet-v1.3} is a large-scale untrimmed video benchmark, contains
19,994 untrimmed videos of 5 to 10 minutes from 200 activity categories.
\end{itemize}

\begin{table*}
\small
\centering
\caption{CLIP-ResNet hyperparameters}
\begin{tabular}{l|ccccccc} \toprule
           & Embedding & Input      & \multicolumn{2}{c}{ResNet}  & \multicolumn{3}{c}{Text Transformer} \\
    Model  & dimension & resolution & blocks & width  & layers & width & heads  \\ \midrule
    RN50   & 1024 & 224 & (3, 4, 6, 3) & 2048 & 12 & 512 & 8 \\
    \bottomrule
\end{tabular}
\label{resnet}
\end{table*}

\begin{table*}[h]
\small
\centering
\caption{CLIP-ViT hyperparameters}
\begin{tabular}{l|cccccccc} \toprule
           & Embedding & Input      & \multicolumn{3}{c}{Vision Transformer} & \multicolumn{3}{c}{Text Transformer} \\
    Model  & dimension & resolution & layers & width & heads  & layers & width & heads \\ \midrule
    ViT-B/32 & 512 & 224 & 12 & 768 & 12 & 12 & 512 & 8 \\
    ViT-B/16 & 512 & 224 & 12 & 768 & 12 & 12 & 512 & 8 \\
    ViT-L/14 & 768 & 224 & 24 & 1024 & 16 & 12 & 768 & 12 \\
    ViT-L/14-336px  & 768 & 336 & 24 & 1024 & 16 & 12 & 768 & 12 \\
    \bottomrule
\end{tabular}
\label{vit}
\end{table*}

\subsection{Large-scale datasets for pre-training}\label{sec:large-scale}
Here we describe the large-scale web-scale datasets used in other video recognition methods for pre-training. The suffix of the name represents the magnitude of the dataset.
\begin{itemize}
    \item \textbf{ImageNet-1K/21K}: The ImageNet-1K dataset was used to pre-train models for computer vision transfer learning. It was first released for the ILSVRC2012 visual recognition challenge. The ImageNet-1K dataset is a subset of the larger ImageNet dataset, which contains 14,197,122 images split into 21,841 categories. The whole dataset is known to as ImageNet-21K (sometimes referred to as ImageNet-22K) and has been \textbf{open-source} \footnote{https://www.image-net.org}. ImageNet-1K was created by selecting a subset of 1.2M images from ImageNet-21K, that belong to 1000 mutually exclusive classes. 
    \item \textbf{IG-65M}: Facebook has proposed the IG-65M dataset, which contains approximately 65 million public, user-generated Instagram videos with hashtags. Due to label and temporal noise, the dataset is used for weakly-supervised training. This dataset is not open-source, but several pre-trained R(2+1)D~\cite{r2+1d} and CSN~\cite{CSN} models are \textbf{provided}~\footnote{https://github.com/facebookresearch/vmz}.
    \item \textbf{JFT-300M}: JFT-300M is an internal Google dataset used to train image classification models. The dataset consists of 300M images that are labeled with 18,291 categories. Image labels are generated using a complex algorithm that combines raw web signals, web page connections, and user feedback. However, the dataset and the pre-trained weights are \textbf{not} open-source.
    \item \textbf{FLD-900M}: FLD-900M is a large image-caption dataset from Microsoft, which includes 900M Images and 900M Free form text (From one word, Phrase to sentence). By now, the dataset and the pre-trained weights are \textbf{not} open-source.
    \item \textbf{JFT-3B}: JFT-3B is an internal Google dataset and a larger version of the JFT-300M. It has over 3 billion images that have been annotated with a class structure of around 30k labels using a semi-automated procedure. Also, the dataset and the pre-trained weights are \textbf{not} open-source.
    \item \textbf{WIT-400M}: WIT-400M is a dataset that contains 400 million web image-text pairs, and is used to train CLIP~\cite{CLIP}. CLIP does not release the dataset, but made all of the pre-trained models \textbf{available} \footnote{https://github.com/openai/CLIP}.
    In this paper, we utilize the CLIP-pretrained models in our experiments.
\end{itemize}

\subsection{Visual encoder architectures}\label{sec:vis_encoder}
We provide the full architecture details of the visual encoder and textual encoders in this paper. Table~\ref{resnet} shows the CLIP-ResNet architectures. Table~\ref{vit} shows the CLIP-ViT architectures.

\begin{algorithm*}[h]
	\caption{Numpy-like Pseudocode that illustrates the role of Batch Gather in Distributed InfoNCE.}
	\label{batch_gather}
	\begin{lstlisting}[language=python]
    # text_encoder: encoder network for text input
    # vision_encoder: encoder network for vision input, e.g., images or videos.
    # V: minibatch of vision inputs
    # T: minibatch of text inputs
    # N: the local batch size of each GPU, e.g.,16
    # M: the number of GPUs, e.g.,8
    # N * M: the global batch size for multi-gpu training, e.g.,128
    
    # extract feature representations of each modality
    local_vision_features = vision_encoder(V) # shape: [N, embed_dim]
    local_text_features = text_encoder(T) # shape: [N, embed_dim]

    # normalization
    local_vision_features = l2_normalize(local_vision_features, axis=1)
    local_text_features = l2_normalize(local_text_features, axis=1)
    
    # batch_gather is a function gathering and concatenating the tensors across GPUs. 
    all_vision_features = batch_gather(local_vision_features) # shape: [N * M, embed_dim]
    all_text_features = batch_gather(local_text_features) # shape: [N * M, embed_dim]
    
    # scaled pairwise cosine similarities
    # shape = [N, N * M]
    logits_per_image = logit_scale * image_features @ all_text_features.t()  
    # shape = [N, N * M]
    logits_per_text = logit_scale * text_features @ all_image_features.t() 
    
    # The logits are then used as inputs for N*M-way (e.g., 128-way) classification, 
    # resulting in a loss value corresponding to N inputs in each GPU. 
    # Then Distributed Data Parallel mechanism takes care of averaging these across GPUs, 
    # which becomes equivalent to calculating the loss over NMxNM (e.g.,128x128) similarities.
	\end{lstlisting}
\end{algorithm*}

\begin{algorithm*}[h]
	\caption{The code generates the LDA coefficient for Kinetics-400 dataset.}
	\label{lda}
	\begin{lstlisting}[language=python]
    import numpy as np
    from sklearn.discriminant_analysis import LinearDiscriminantAnalysis as LDA
    input = np.load('feats_labels_400class.npz')   # pre-extracted visual features
    feats = input['feats']  # size: [24000, 512]
    labels = input['labels']  # size: [24000,]
    lda = LDA()
    lda.fit(feats, labels)
    classifier = lda.coef_ # size: [400, 512]	
	\end{lstlisting}
\end{algorithm*}

\subsection{Batch Gather for Distributed InfoNCE}\label{sec:batch_gather}
\label{sm:batch_gather}
Instead of Data-Parallel Training (DP), which is single-process, multi-thread, and only works on a single machine, Distributed Data-Parallel Training (DDP) is a widely adopted single-program multiple-data training paradigm for single- and multi-machine training.
Due to GIL contention across threads, per-iteration replicated model, and additional overhead introduced by scattering inputs and gathering outputs, DP is usually slower than DDP even on a single machine.
Hence, we develop the Distributed InfoNCE based on DDP for large batch size and fast training.
The core of the Distributed InfoNCE implementation is batch gathering.
Say there are M GPUs and each GPU gets N input pairs, we need to calculate the NM\x NM similarity matrix across the GPUs for InfoNCE loss. 
Without batch gathering, each GPU only computes a local N\x N matrix, \emph{s.t.} N$\ll$NM, Then the cosine similarity and the InfoNCE loss would be calculated only for the pairs within a single GPU and later their gradients would be averaged and synced. 
That's obviously not what we want.

The batch gathering for Distributed InfoNCE is presented as follows.
When calculating the similarity matrix (and thus the logit scores across text inputs for each image/video), a GPU only needs to hold M vision features, and perform matrix product with NM text features, yielding an M\x NM matrix. This computation is distributed (\emph{i.e.}, sharded) across N GPUs, and we have calculated NM\x NM similarities across the GPUs in total. The loss we employ is symmetric and the same happens \emph{w.r.t.} text inputs.
As shown in \Cref{batch_gather}, we also give an example pseudocode to help you understand the statement.

\subsection{LDA classifier}\label{supp:lda}
Here we provide the details of LDA classifier.
We directly use the official CLIP-pretrained visual encoder to extract video embeddings, and the visual encoder is not finetuned on Kinetics-400. Then we perform LDA on the pre-extracted video embeddings of the training set in Kinetics-400 to initialize W and freeze it for finetuning the visual encoder on the Kinetics-400 dataset.

LDA is commonly used for feature classification or feature dimensionality reduction. However, in this work, we only use LDA for feature classification (in order to get ``discriminant coefficients" as the classifier) instead of feature dimensionality reduction.
For better understanding, we show the code in \Cref{lda} which generates the LDA coefficient and there is no dimension reduction.

\subsection{Discussion on data overlaps}\label{sec:overlaps}
In this paper, we mainly focus on the video recognition task with the Kinetics dataset. As shown in Fig.17 of CLIP official paper, CLIP has done the data overlap analysis on the Kinetics-700 dataset. They observe that there are less than 1\% overlaps and many overlaps on Kinetics-700 are in fact all black transition frames. Then they conduct the experiment on overlapping data. The results show that the Kinetics-700 has no performance improvement, and even has an apparent 20\% accuracy drop on the overlapping data. Therefore, the use of data in this paper is reasonable.

\section{Additional Results}\label{sec:results}
\subsection{Comparison with state-of-the-arts on UCF-101 and HMDB-51}\label{sec:small_sota}
We also evaluate our method on the UCF-101 and HMDB-51 datasets to demonstrate its capacity to generalize to smaller datasets.
We finetune our models on these two datasets using the pre-trained ViT-L model on Kinetics-400 and present the mean class accuracy on split one. We utilize 16 frames as inputs and 30 epochs for training.
Table \ref{t:UCFHMDB} reveals that our model has a pretty transfer capability, with mean class accuracy of 98.2\% on UCF-101 and 81.3\% on HMDB-51, respectively.

\begin{table}[h]
\centering
\footnotesize
\begin{tabular}{lcc}
\toprule
\textbf{Method}   & \textbf{UCF-101}  & \textbf{HMDB-51} \\ \midrule

ARTNet~\cite{ARN}  & 94.3\% & 70.9\% \\  
I3D~\cite{i3d}    & 95.6\%  & 74.8\% \\ 
R(2+1)D~\cite{r2+1d}    & 96.8\%  & 74.5\% \\
S3D-G~\cite{s3d} &  96.8\%  & 75.9\% \\ 
TSM~\cite{tsm} &  95.9\%  & 73.5\%  \\
STM~\cite{stm}  &  96.2\%  & 72.2\%   \\ 
TEINet~\cite{teinet}  &  96.7\%  & 72.1\%   \\ 
MVFNet~\cite{wu2020MVFNet}  &  96.6\%  & 75.7\%  \\ 
TDN~\cite{tdn} & 97.4\% & 76.4\% \\ \midrule
Ours ViT-L & \textbf{98.1\%} & \textbf{81.3}\% \\
Ours ViT-L (336$\uparrow$) & \textbf{98.2\%} & \textbf{81.3}\% \\
\bottomrule
\end{tabular}
\caption{\textbf{Mean class accuracy} on UCF-101 and HMDB-51 achieved by different methods which are transferred from their \textbf{Kinetics} models with RGB modality.}
\label{t:UCFHMDB}
\end{table}

\subsection{More visualizations of different classifiers}\label{sec:vis}
\begin{figure*}
\begin{center}
\includegraphics[width=0.9\textwidth]{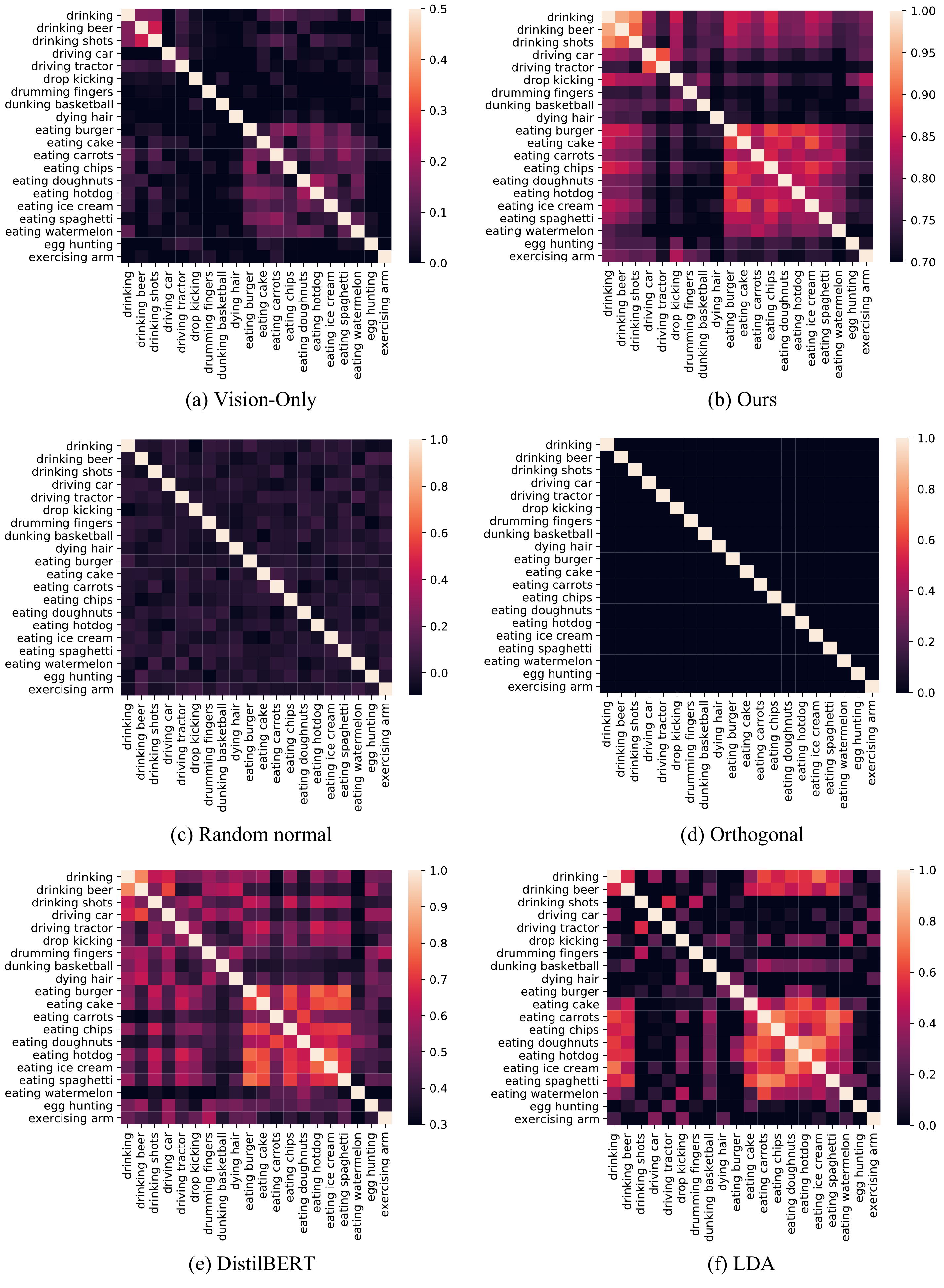}
\end{center}
\caption{Inter-class correlation maps of ``embeddings of class labels'' for 20 categories on Kinetics-400. The color thresholds are adjusted for better understandability. Please zoom in for best view.}
\label{fig:heatmap}
\end{figure*}

Here we provide more visualizations of different classifiers in Figure~\ref{fig:heatmap}.

\begin{figure*}[t] 
  \begin{minipage}[]{0.48\textwidth} 
    \centering 
    \includegraphics[width=0.95\linewidth]{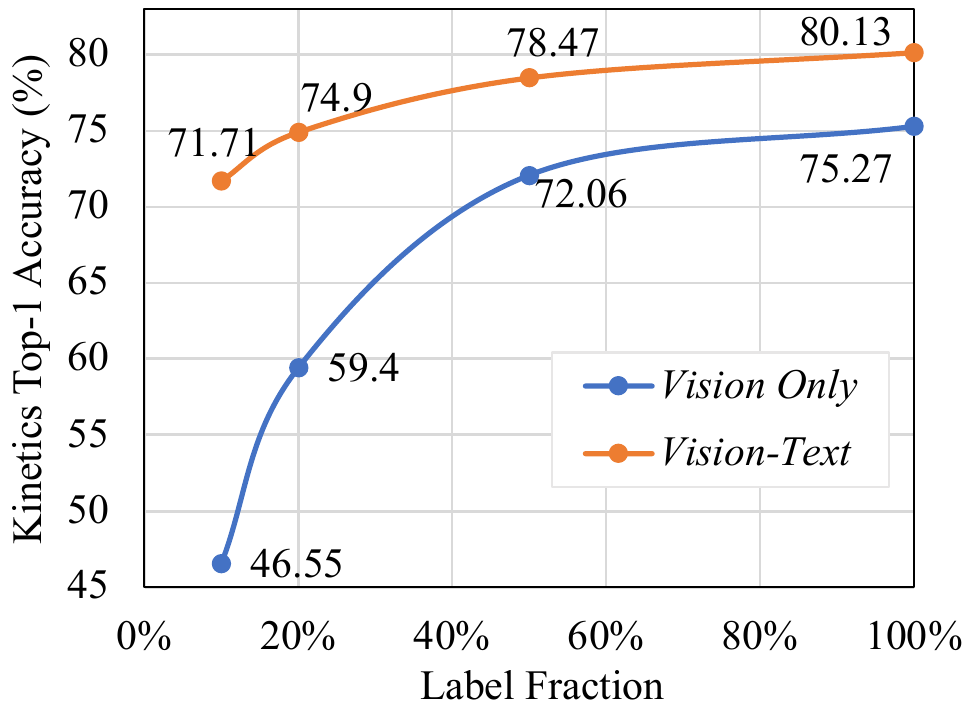} 
    \caption{Vision-Text \vs Vision-only framework under different label fractions on Kinetics-400.} 
    \label{fig:label fraction}
  \end{minipage}%
  \quad 
  \begin{minipage}[]{0.48\textwidth} 
    \centering 
    \includegraphics[width=0.95\linewidth]{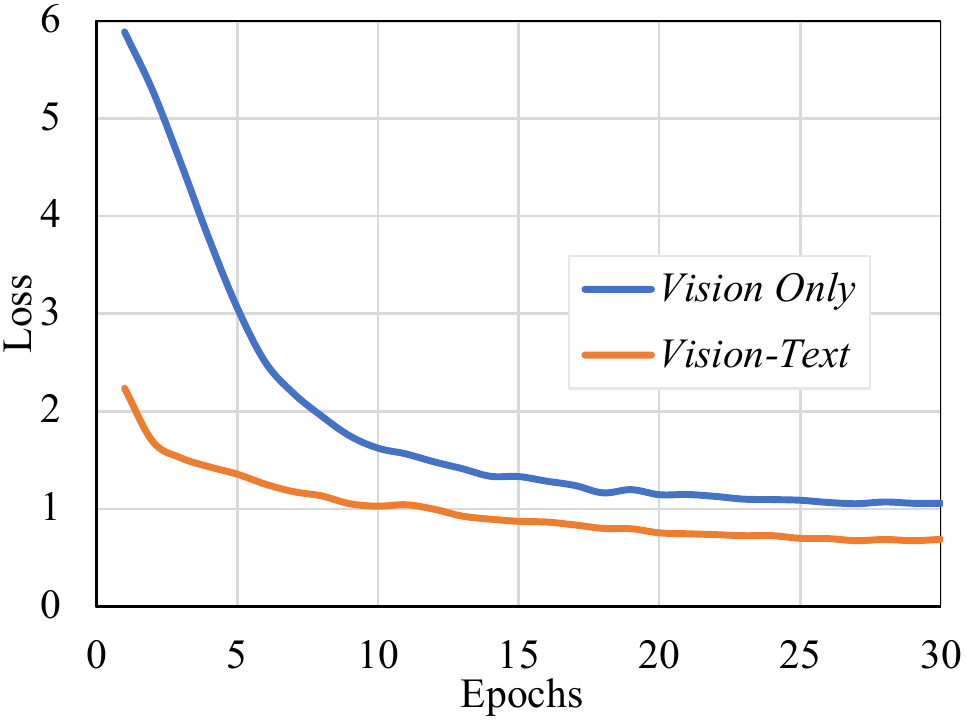} 
    \caption{The training loss of Vision-Text and Vision-only framework on Kinetics-400.} 
    \label{fig:loss}
  \end{minipage}
\end{figure*}

\subsection{More ablations on Kinetics-400.}\label{sec:ab}

\textbf{Comparison with vision-only framework.}
Here we present additional comparison figures.
As presented in Figure~\ref{fig:label fraction}, our method surpasses \emph{Vision-Only} baselines across multiple label fractions on Kinetics-400.
Especially when just only 10\% labeled data is available for training, demonstrating that the advantage of our paradigm is more profound when the labeled data is limited.
Also, when training with full data, our \emph{Vision-Text} method leads to an additional 5\% improvement with the same training recipe.
Figure~\ref{fig:loss} further demonstrates our paradigm significantly improves convergence speed.



\textbf{Text input forms.}
We study several text input forms in Table~\ref{table:text input}, including class names, single hard template, multiple hard templates, and learnable templates. More details are as follows:
\begin{itemize}
    \item \textbf{Class name} To build textual embeddings, we utilize the category names of the dataset as the text input, \eg, \emph{``eating hotdog''}, \emph{``driving car''}, \emph{etc}. We can see that only using the label text can yield good results
    \item \textbf{Single hard template} We employ the hand-crafted template \emph{``a video of a person \{class name\}.''} to form a sentence as input. This only slightly increases performance over the baseline of using the label text.
    \item \textbf{Multiple hard templates} CLIP~\footnote{https://github.com/openai/CLIP/blob/main/data/prompts.md} provides 28 templates for Kinetics, one of which is the above single template.
We use these multiple templates as the text augmentation during training.
At each iteration, we choose one template at random as text input. 
Then, using the above single hard template as input, we perform the evaluation.
Performance decreases by 0.64\% on Kinetics-400. This may be because different prompt templates may introduce extra noise for the training.
    \item \textbf{Learnable templates}
We adopt the automated prompt CoOp~\cite{coop} to describe a prompt's context using a set of learnable vectors. 
Specifically, the prompt given to the text encoder is designed with the following form,
\begin{equation} \label{eq:prompt_cls_end}
\bm{t} = [\text{V}]_1 [\text{V}]_2 \hdots [\text{V}]_M [\text{class name}],
\end{equation}
where each $[\text{V}]_m$ ($m\!\in\!\{1, \hdots, M\}$) is a vector of the same size as word embeddings, and $M$ is a hyperparameter indicating the number of context tokens. We set the $M$ to 4.
The results suggest that different templates have little impact on our model.
\end{itemize}

\begin{table}[h]
    \centering
  \begin{tabular}{lc}
  \toprule
  Text input form & Top-1\\
  \midrule
  class name & 81.4  \\
  ``a video of a person'' + class name  & \baseline{\textbf{81.5}}  \\
  multiple fixed templates + class name & 80.9 \\
  learnable template + class name & 81.2  \\ \bottomrule
   \end{tabular}
   \caption{Study on various text input forms.}
  \label{table:text input}
\end{table}






\end{document}